%% file: iclr2026_conference.tex
\documentclass{article} 
\usepackage{iclr2026_conference,times}

\input{math_commands.tex}

\usepackage{hyperref}
\usepackage{url}

\title{MoRA: Missing Modality Low-Rank Adaptation for Visual Recognition}

\author{%
  Shu Zhao\textsuperscript{\rm 1}, Nilesh Ahuja\textsuperscript{\rm 2}, Tan Yu\textsuperscript{\rm 3}, Tianyi Shen\textsuperscript{\rm 1}, Vijaykrishnan Narayanan\textsuperscript{\rm 1} \\
  \textsuperscript{\rm 1}The Pennsylvania State University 
  \textsuperscript{\rm 2}Intel \textsuperscript{\rm 3}NVIDIA\\
  \{smz5505, vijaykrishnan.narayanan\}@psu.edu
}

\input{sections/user_defined}

\iclrfinalcopy
\begin{document}

\maketitle

\input{sections/0_abstract}
\input{sections/1_introduction}
\input{sections/2_related_work}
\input{sections/3_method}
\input{sections/4_experiments}
\input{sections/5_conclusion}

\section{Acknowledgments}
This work was supported in part by Semiconductor Research Corporation JUMP 2.0 PRISM Center.

\bibliography{iclr2026_conference}
\bibliographystyle{iclr2026_conference}

\input{sections/appendix}

\end{document}

%% file: math_commands.tex
\usepackage{amsmath,amsfonts,bm}

\def\eqref#1{equation~\ref{#1}}

\def\1{\bm{1}}

\DeclareMathAlphabet{\mathsfit}{\encodingdefault}{\sfdefault}{m}{sl}
\SetMathAlphabet{\mathsfit}{bold}{\encodingdefault}{\sfdefault}{bx}{n}



%% file: sections/user_defined.tex
\usepackage{booktabs}
\usepackage{multirow}
\usepackage{xcolor}
\usepackage{colortbl}
\usepackage{graphicx}
\usepackage{amsmath}
\usepackage{cleveref}
\usepackage{cancel}

\usepackage{caption}
\usepackage{float}

%% file: sections/0_abstract.tex
\begin{abstract}
    Pre-trained vision language models have shown remarkable performance on visual recognition tasks, but they typically assume the availability of complete multimodal inputs during both training and inference. In real-world scenarios, however, modalities may be missing due to privacy constraints, collection difficulties, or resource limitations. While previous approaches have addressed this challenge using prompt learning techniques, they fail to capture the cross-modal relationships necessary for effective multimodal visual recognition and suffer from inevitable computational overhead. In this paper, we introduce MoRA, a parameter-efficient fine-tuning method that explicitly models cross-modal interactions while maintaining modality-specific adaptations. MoRA introduces modality-common parameters between text and vision encoders, enabling bidirectional knowledge transfer. Additionally, combined with the modality-specific parameters, MoRA allows the backbone model to maintain inter-modality interaction and enable intra-modality flexibility. Extensive experiments on standard benchmarks demonstrate that MoRA achieves an average performance improvement in missing-modality scenarios by $5.24\%$ and uses only $25.90\%$ of the inference time compared to the SOTA method while requiring only $0.11\%$ of trainable parameters compared to full fine-tuning. The code is available at \url{https://github.com/Tree-Shu-Zhao/MoRA}.
\end{abstract}

%% file: sections/1_introduction.tex
\section{Introduction}
Pre-trained vision language models~(VLMs) integrate multiple modalities~(e.g., vision and language) to comprehensively understand their environment, demonstrating remarkable performance on various downstream tasks, including visual recognition~\citep{DCP_NeurIPS24}, cross-modal retrieval~\citep{tac_iclr25}, and retrieval-augmented generation~\citep{zhao2025windsock}. VLMs like CLIP~\citep{CLIP_ICML21} and ViLT~\citep{VILT_ICML21} leverage large-scale paired data to learn joint representations of images and text. Multimodal large language models, including GPT-4~\citep{gpt4_arxiv23}, Gemini~\citep{gemini_arxiv24}, LLaMA-Vision~\citep{llama3_arxiv24}, and LLaVA~\citep{llava_neurips23}, build connections between vision and language and use the knowledge within LLMs to establish powerful conversation and reasoning abilities.

Despite their impressive capabilities, deploying them in real-world scenarios presents two significant challenges. First, most multimodal models work under the assumption of modality completeness~\citep{neucore}, requiring all modalities to be available during both training and inference. However, this assumption rarely holds in practice due to privacy constraints, collection difficulties, or resource limitations~\citep{first_mmt_CVPR22,hao2025safemap}. When input modalities are missing, performance degrades substantially~\citep{DCP_NeurIPS24}, limiting their applicability in real-world settings where data completeness cannot be guaranteed. Second, as model sizes grow, fine-tuning becomes increasingly computationally expensive with limited resources and leads to overfitting on small-scale target datasets~\citep{MaPLe_CVPR23}. Although several works~\citep{MMP_CVPR23,DCP_NeurIPS24} have devised prompt-based methods to alleviate them, the prompts lead to inevitable inference overhead.

To address these challenges, we explore the underlying mechanisms affecting the performance when modalities are missing. A critical insight comes from Mind the Gap~\citep{MindTheGap_NeurIPS22}, identifying the ``modality gap'' which is the geometric separation between different modality embeddings in the shared representation space. Building on this observation, we argue that both the alignment and gap between modalities provide valuable complementary information for improving performance during inference with missing modalities. Specifically, during fine-tuning, the embedding spaces of the visual and text encoders should be related, moving in the same direction to maintain multimodal performance. Simultaneously, these encoders need to maintain their own independent update directions to better adapt to downstream tasks without compromising modality-specific characteristics.

Inspired by this, we introduce MoRA, a parameter-efficient fine-tuning method that explicitly models cross-modal interactions while maintaining modality-specific adaptations. MoRA incorporates two key design elements: a shared cross-modal parameter module that enables knowledge transfer between modalities through the Gram matrix~\citep{strang2022introduction} of shared low-rank parameters and modality-specific adaptation components that preserve the unique characteristics of each modality. This dual-structure design allows MoRA to maintain inter-modality interactions while enabling intra-modality flexibility, resulting in robust performance across various missing-modality scenarios.

To summarize our contributions, we propose MoRA, a parameter-efficient fine-tuning method for multimodal models that explicitly addresses the challenge of missing modalities through shared cross-modal parameters and modality-specific adaptations, enabling bidirectional knowledge transfer between modalities while preserving the directional properties of the original weights. We design an efficient training strategy that requires updating only a small fraction ($\sim0.11\%$) of the model parameters, making it feasible to adapt large pre-trained models even with limited computational resources. Through extensive experiments on standard benchmarks, we demonstrate that MoRA significantly outperforms existing prompt-based and parameter-efficient approaches across various missing-modality scenarios while maintaining inference efficiency.

%% file: sections/2_related_work.tex
\section{Related Work}
\subsection{Missing Modality for Multimodal Learning}
The missing modality issue presents a significant challenge in deploying robust systems, leading to a significant performance drop. Previous approaches for addressing missing modality challenges can be broadly categorized into Alignment-based and Reconstruction-based methods. Alignment-based methods~\citep{ShaSpec_CVPR23,UnseenInteraction_NeurIPS23,EverythingAtOnce_CVPR22} embed different modalities into a shared representation space, enabling the model to operate effectively even when certain modalities are missing by aligning the feature spaces of different modalities during pre-training or fine-tuning. Reconstruction-based methods~\citep{first_mmt_CVPR22,MMIN_ACL21,SMIL_AAAI21} use available modalities to reconstruct features of missing modalities explicitly. These approaches typically employ generative models or cross-modal translation networks to synthesize the absent information, allowing the model to operate on ``completed'' inputs. However, these methods often suffer from imperfect reconstruction quality, especially when the missing modality contains information that cannot be fully inferred from available ones. More recently, prompt learning techniques~\citep{MMP_CVPR23,DCP_NeurIPS24} have emerged as a subset of reconstruction-based approaches, handling missing-modality scenarios by inserting learnable tokens into transformer layers. Modality-specific information is offloaded to learnable prompts and reused when modalities are missing. MMP~\citep{MMP_CVPR23} treats different missing-modality cases as different types of input, adapting the model through learnable prompts while keeping the backbone frozen. However, MMP inserts independent prompts into each layer, overlooking the complex relationships between modalities. DCP~\citep{DCP_NeurIPS24} and SyP~\citep{SYP_arXiv25} devise more prompts to leverage the correlations between prompts and input features across different layers. However, it discards the features of the missing modalities and cannot fully exploit  multimodal features for downstream tasks. MoRA preserves the modality information during training and introduces no overhead during inference.

\subsection{Parameter-Efficient Fine-Tuning}
Large language models have been widely used in various downstream tasks~\citep{parallelsearch,expandsearch}. Parameter-Efficient Fine-Tuning~(PEFT) methods reduce the computational burden of adapting large models by updating only a small subset of parameters. These approaches can be classified into three categories. Adapter-based methods~\citep{Adapter_ICML19} insert trainable modules into backbones, either sequentially or in parallel with existing layers. Prompt-based methods~\citep{PromptSurvey23} add trainable tokens to the input while keeping model parameters fixed. Both categories typically introduce additional inference latency. Low-Rank Adaptation~(LoRA) methods~\citep{LoRA_ICLR22} approximate weight updates using low-rank matrices that can be merged with pre-trained weights before inference, thus maintaining inference efficiency. Various extensions have been proposed, including SVD-based approaches~\citep{AdaLoRA_ICLR23}, orthogonal factorization~\citep{DBLP:conf/nips/QiuLFXFLZWS23,DBLP:conf/iclr/LiuQFXXYF0HPWBW24}, and direction decomposition~\citep{DoRA_ICML24}. While Multimodal LoRA methods~\citep{mmlora1,mmlora2} have focused on instruction tuning, they cannot handle missing modalities and lack architectural innovations for cross-modal interaction in dual-branch architectures. \citet{DBLP:conf/miccai/ShiKLLP24} address the task in medical diagnosis through unidirectional adaptation. MoRA targets general visual recognition tasks with bidirectional knowledge transfer, achieving superior efficiency with smaller trainable parameters and zero inference latency.

%% file: sections/3_method.tex
\section{Method}
\subsection{Problem Formulation}
We focus on the multimodal classification task with missing modalities during both training and testing. For simplicity, but without loss of generality, we consider a multimodal dataset with text~($\mathrm{t}$) and vision~($\mathrm{v}$) modalities, i.e., $\mathcal{D} = \{\mathcal{D}^{\mathrm{t}}, \mathcal{D}^{\mathrm{v}}, \mathcal{D}^{\mathrm{c}}\}$. Specifically, $\mathcal{D}^{\mathrm{t}} = \{(\mathbf{t}_i, \mathbf{y}_i)\}_{i=1}^{N_{\mathrm{t}}}$ contains text-only data samples; $\mathcal{D}^{\mathrm{v}} = \{(\mathbf{v}_i, \mathbf{y}_i)\}_{i=1}^{N_{\mathrm{v}}}$ includes image-only data samples; $\mathcal{D}^{\mathrm{c}} = \{(\mathbf{t}_i, \mathbf{v}_i, \mathbf{y}_i)\}_{i=1}^{N_{\mathrm{c}}}$ is the subset containing modality-complete samples with both text and image, where $\mathbf{t}_i$ is text, $\mathbf{v}_i$ denotes an image, and $\mathbf{y}_i \in \mathbb{R}^{C}$ is the label vector where $C$ is the number of classes. When the image is missing, we set the image input to an all-$1$ matrix; when the text is missing, we set the text input to an empty string.

\input{figures/fig_motivation}
\input{figures/fig_mora}

\subsection{Motivation} \label{sec:motivation}

Vision Language Models~(VLMs) have been pre-trained on massive image-text pairs. Although the pre-training stage aligns the vision and language embedding space, Mind the Gap~\citep{MindTheGap_NeurIPS22} points out that there is still a gap between modalities. We argue that both the alignment property and the gap are important for the missing modality task. To demonstrate these, we fine-tune aligned and unaligned models using modality-complete samples and test them using both complete and incomplete samples, as illustrated in \Cref{fig:motivation} (a) (b). The aligned/unaligned models denote whether vision and text encoders are trained on image-text pairs. Implementation details can be found in \Cref{sec:implementation_details}. Compared to the performance drop of $-11.1$ using the aligned model, the unaligned model shows a drop of $-54.5$, demonstrating that other available aligned modalities can maintain a certain level of performance when modalities are missing. Additionally, using both image and text features with the aligned model achieves better performance than using only image features ($-14.1$) or only text features ($-11.9$). This finding suggests that the gap represents different information across modalities, which serves as important complementary information for multimodal tasks.

Therefore, we identify two properties that need to be considered during fine-tuning VLMs, as illustrated in \Cref{fig:motivation}~(c). First, the direction of fine-tuning image and text modalities should be the same to maintain their relationship in embedding space for general ability. Second, the image and text modalities should have their own fine-tuning direction to enable flexibility for downstream tasks. Inspired by these, we propose MoRA, a parameter-efficient fine-tuning method that explicitly models cross-modal interactions while maintaining modality-specific adaptations.

\subsection{Missing Modality Low-Rank Adaptation} \label{sec:mora}
The weight matrix in LoRA can be decomposed into the magnitude and direction, as shown below:
\begin{equation} \label{eq:dora}
    \mathbf{W} = \mathbf{W}_0 + \Delta \mathbf{W} = \Vert \mathbf{W}_0 + \Delta \mathbf{W} \Vert_\mathrm{F} \frac{\mathbf{W}_0 + \Delta \mathbf{W}}{\Vert \mathbf{W}_0 + \Delta \mathbf{W} \Vert_\mathrm{F}} = \Vert \mathbf{W}_0 + \Delta \mathbf{W} \Vert_\mathrm{F} \overline{\mathbf{W}_0 + \Delta \mathbf{W}},
\end{equation}
where $\Vert \mathbf{W} \Vert_{\mathrm{F}}$ is the Frobenius norm of the matrix, denoting the magnitude; $\overline{\mathbf{W}}$ is the normalized matrix, denoting the direction.

Although recent works~\citep{DoRA_ICML24,sdlora_ICLR25} have shown the importance of the direction in fine-tuning models, they focus on large language models, while the cross-modality information interaction, which is important for multimodal tasks, is not discussed. Based on the analysis in \Cref{sec:motivation}, we introduce MoRA, a parameter-efficient fine-tuning
method that enables cross-modal interactions and captures modality-common/specific information during training, as illustrated in \Cref{fig:mora}. MoRA introduces two types of learnable parameters, including modality-specific parameters $\mathbf{A}^{\mathrm{v}/\mathrm{t}} \in \mathbb{R}^{r \times d_{\mathrm{v}/\mathrm{t}}}, \mathbf{B}^{\mathrm{v}/\mathrm{t}} \in \mathbb{R}^{d_{\mathrm{v}/\mathrm{t}} \times r}$ for independent adaptation, and shared parameters $\mathbf{S}^v \in \mathbb{R}^{r \times d_{\mathrm{v}}}, \mathbf{S}^t \in \mathbb{R}^{d_{\mathrm{t}} \times r}$ for cross-modal knowledge transfer, where $d_{\mathrm{v}}$ and $d_{\mathrm{t}}$ are the dimensions of vision and text encoders respectively, and $r \ll d$ is the rank. The updated weight matrix for image~($\mathrm{v}$) / text~($\mathrm{t})$ encoders is:
\begin{equation} \label{eq:mora_ori}
\begin{aligned}
    \mathbf{W}^{\mathrm{v}/\mathrm{t}} &= \mathbf{W}_0^{\mathrm{v}/\mathrm{t}} + \Delta \mathbf{W}^{\mathrm{v}/\mathrm{t}} + \Delta \mathbf{W}^{\mathrm{s}} \\
    &= \left(\mathbf{W}_0^{\mathrm{v}/\mathrm{t}} + \Delta \mathbf{W}^{\mathrm{v}/\mathrm{t}}\right) + \left(\mathbf{W}_0^{\mathrm{v}/\mathrm{t}} + \Delta \mathbf{W}^{\mathrm{s}}\right) \cancel{- \mathbf{W}_0^{\mathrm{v}/\mathrm{t}}} \quad \text{$-\mathbf{W}_0^{\mathrm{v}/\mathrm{t}}$ is frozen and ignored} \\ 
    &= \Vert \mathbf{W}_0^{\mathrm{v}/\mathrm{t}} + \Delta \mathbf{W}^{\mathrm{v}/\mathrm{t}} \Vert_{\mathrm{F}}\frac{\mathbf{W}_0^{\mathrm{v}/\mathrm{t}} + \Delta \mathbf{W}^{\mathrm{v}/\mathrm{t}}}{\Vert \mathbf{W}_0^{\mathrm{v}/\mathrm{t}} + \Delta \mathbf{W}^{\mathrm{v}/\mathrm{t}} \Vert_{\mathrm{F}}} + \Vert \mathbf{W}_0^{\mathrm{v}/\mathrm{t}} + \Delta \mathbf{W}^{\mathrm{s}} \Vert_{\mathrm{F}}\frac{\mathbf{W}_0^{\mathrm{v}/\mathrm{t}} + \Delta \mathbf{W}^{\mathrm{s}}}{\Vert \mathbf{W}_0^{\mathrm{v}/\mathrm{t}} + \Delta \mathbf{W}^{\mathrm{s}} \Vert_{\mathrm{F}}} \\
    &= \Vert \mathbf{W}_0^{\mathrm{v}/\mathrm{t}} + \Delta \mathbf{W}^{\mathrm{v}/\mathrm{t}} \Vert_{\mathrm{F}} \overline{\mathbf{W}_0^{\mathrm{v}/\mathrm{t}} + \Delta \mathbf{W}^{\mathrm{v}/\mathrm{t}}} + \Vert \mathbf{W}_0^{\mathrm{v}/\mathrm{t}} + \Delta \mathbf{W}^{\mathrm{s}} \Vert_{\mathrm{F}} \overline{\mathbf{W}_0^{\mathrm{v}/\mathrm{t}} + \Delta \mathbf{W}^{\mathrm{s}}} \\
    &=\underbrace{\alpha^{\mathrm{v}/\mathrm{t}}\overline{\mathbf{W}_0^{\mathrm{v}/\mathrm{t}} +  \mathbf{B}^{\mathrm{v}/\mathrm{t}}{\mathbf{A}^{\mathrm{v}/\mathrm{t}}}}}_{\text{Modality-Specific}} + \underbrace{\alpha^{\mathrm{s}}\overline{\mathbf{W}_0^{\mathrm{v}/\mathrm{t}} + \mathbf{S}^{\mathrm{v}/\mathrm{t}}\mathbf{S}^{\mathrm{t}/\mathrm{v}}}}_{\text{Modality-Shared}},
\end{aligned}
\end{equation}
where $\alpha^{\mathrm{v}/\mathrm{t}}$ and $\alpha^{\mathrm{s}}$ denote the learnable modality-specific and modality-shared magnitudes; $\mathbf{W}_0^{\mathrm{v}/\mathrm{t}}$ is the frozen pre-trained weights in vision/text encoders.

However, \Cref{eq:mora_ori} only works when the dimensions of the image and text encoders are the same. For example, the dimension of vision ($d_\mathrm{v}$) and text encoders ($d_\mathrm{t})$ in the CLIP ViT-B/16 model is $768$ and $512$, i.e., $\mathbf{W}_0^{\mathrm{v}} \in \mathbb{R}^{768 \times 768}$ and $\mathbf{W}_0^{\mathrm{t}} \in \mathbb{R}^{512 \times 512}$. Direct multiplication of $\mathbf{S}^\mathrm{v}$ and $\mathbf{S}^\mathrm{t}$ would yield $\mathbb{R}^{r \times d_\mathrm{v}} \times \mathbb{R}^{r \times d_\mathrm{t}}$, which is incompatible with both encoder dimensions. This dimension mismatch challenge significantly limits the applicability of MoRA. Although we can add projection layers to map the image and text embeddings to a common space, the projection layers will significantly increase the number of learnable parameters during training and cannot be absorbed into the pre-trained weights $\mathbf{W}_0$, increasing the inference latency. We resolve this dimension mismatch by operating in the rank space through Gram matrices~\citep{strang2022introduction}. For shared parameters $\mathbf{S}^\mathrm{v}$ and $\mathbf{S}^\mathrm{t}$, we compute:
\begin{equation}
    \begin{aligned}
        \mathbf{G}^{\mathrm{v}} &= \mathbf{S}^{\mathrm{v}} {\mathbf{S}^{\mathrm{v}}}^T \in \mathbb{R}^{r \times r} \\
        \mathbf{G}^{\mathrm{t}} &= \mathbf{S}^{\mathrm{t}} {\mathbf{S}^{\mathrm{t}}}^T \in \mathbb{R}^{r \times r}.
    \end{aligned}
\end{equation}
The key insight is that these Gram matrices capture the structural information of each modality in a dimension-agnostic rank space. We then use cross-modal Gram matrices to update each encoder:
\begin{equation}
    \begin{aligned}
        \mathbf{W}^{\mathrm{v}} &= \alpha^{\mathrm{v}}\overline{\mathbf{W}_0^{\mathrm{v}} + \mathbf{B}^{\mathrm{v}}{\mathbf{A}^{\mathrm{v}}}} + \alpha^{\mathrm{s}}\overline{\mathbf{W}_0^{\mathrm{v}} + {{\mathbf{S}^{\mathrm{v}}}^T}\mathbf{G}^{\mathrm{t}}\mathbf{S}^{\mathrm{v}}} \\
        \mathbf{W}^{\mathrm{t}} &= \alpha^{\mathrm{t}}\overline{\mathbf{W}_0^{\mathrm{t}} + \mathbf{B}^{\mathrm{t}}{\mathbf{A}^{\mathrm{t}}}} + \alpha^{\mathrm{s}}\overline{\mathbf{W}_0^{\mathrm{t}} + {{\mathbf{S}^{\mathrm{t}}}^T}\mathbf{G}^{\mathrm{v}}\mathbf{S}^{\mathrm{t}}}.
    \end{aligned}
\end{equation}
Since ${{\mathbf{S}^{\mathrm{v}}}^T}\mathbf{G}^{\mathrm{t}}\mathbf{S}^{\mathrm{v}} \in \mathbb{R}^{d_{\mathrm{v}} \times d_{\mathrm{v}}}$ and ${{\mathbf{S}^{\mathrm{t}}}^T}\mathbf{G}^{\mathrm{v}}\mathbf{S}^{\mathrm{t}} \in \mathbb{R}^{d_{\mathrm{t}} \times d_{\mathrm{t}}}$, they can be absorbed into the pre-trained weights during inference.

\textbf{Discussion}\quad First, the rank space captures second-order statistics of the low-rank representations, which extracting invariant representations across domains~\citep{arjovsky2019invariant}.
Second, the low-rank structure serves as a cross-modal adaptation module that transforms modality-specific parameters to incorporate shared knowledge~\citep{srebro2005rank}. With Gram matrices, MoRA maintains a balance between preserving modality-specific characteristics and enabling cross-modal information exchange. More importantly, all introduced learnable parameters can be absorbed into the original pre-trained weights, which makes \textbf{MoRA introduce no overheads during inference}.

%% file: figures/fig_motivation.tex
\begin{figure}[t]
    \centering
    \includegraphics[width=0.8\linewidth]{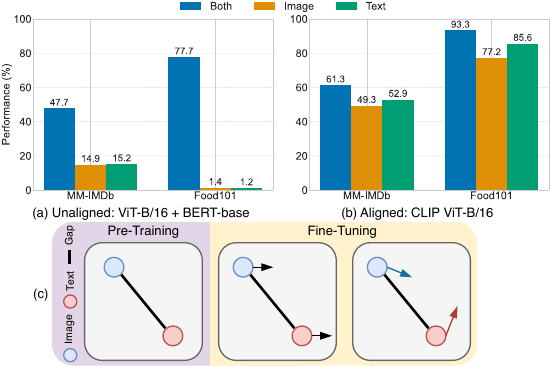}
    \caption{Motivation for MoRA. (a) Performance comparison on MM-IMDb and Food101 datasets using unaligned vision and text encoders. (b) Performance comparison using aligned CLIP ViT-B/16 encoder. (c) During pre-training, modalities are aligned in embedding space with a gap; during fine-tuning, modalities should maintain their relationship while allowing modality-specific adaptations.}
    \label{fig:motivation}
\end{figure}

%% file: figures/fig_mora.tex
\begin{figure}[t]
    \centering
    \includegraphics[width=0.9\linewidth]{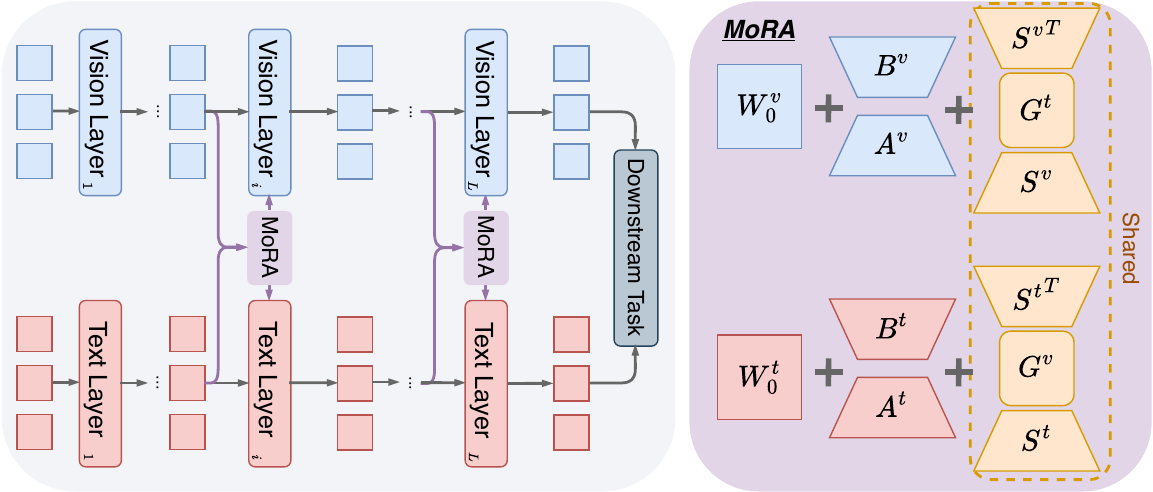}
    \caption{Overview of the proposed MoRA architecture.}
    \label{fig:mora}
\end{figure}

%% file: sections/4_experiments.tex
\section{Experiments}

\subsection{Experimental Setup} \label{sec:experiment_setup}
We evaluate MoRA on three benchmarks, including MM-IMDb~\citep{mmimdb_iclrw17}, UPMC-Food101~\citep{food101_icmew15}, and Hateful Memes~\citep{hatefulmemes_neurips20}. We adopt F1-Macro, top-$1$ classification accuracy, and Area Under the Receiver Operating Characteristic Curve (AUROC) to evaluate the three benchmarks, respectively. More details can be found in \Cref{sec:experiment_setup_details}.

\textbf{Missing Modality Setting}\quad We adopt a rigorous approach wherein modality absence occurs throughout both the training and inference phases. Following previous works~ \citep{MMP_CVPR23,DCP_NeurIPS24}, we designate $\eta\%$ as the missing ratio that quantifies the proportion of incomplete-modality data. In single-modality missing scenarios, the distribution follows a ratio of $\eta\%$ incomplete-modality samples to $1-\eta\%$ complete-modality samples. When addressing dual-modality absences, the dataset consists of $\frac{\eta}{2}\%$ image-only instances and $\frac{\eta}{2}\%$ text-only instances, complemented by $1-\eta\%$ of samples containing both modalities. This configuration effectively simulates real-world modality scarcity conditions and provides a robust framework for evaluating performance in missing modality environments. Implementation details can be found in \Cref{sec:experiment_setup_details}.

\input{tables/tab_main}

\subsection{Main Results}
As shown in \Cref{tab:main}, MoRA consistently outperforms baseline methods across all missing ratio settings. Several key findings emerge from our experiments. First, the text modality consistently demonstrates greater importance than the image modality across all three datasets. This asymmetry may partially stem from text containing direct label information in certain datasets like UPMC-Food101. Second, MoRA achieves particularly strong improvements when the image modality is missing, highlighting its effectiveness in addressing the inadequate visual understanding of current methods through cross-modal knowledge interaction. Third, MoRA maintains remarkable robustness even under extreme conditions with a $90\%$ missing ratio on Hateful Memes, it achieves performance comparable to DCP at only $50\%$ missing ratio, demonstrating its superior ability to handle severe modality scarcity. These results validate that our dual mechanism of modality-specific adaptation and cross-modal parameter sharing creates a more resilient multimodal learning framework.

\subsection{Cross-Scenario Generalization}
\input{figures/fig_generalizability_hatememes}
To evaluate the generalization capability of MoRA across different missing-modality scenarios, we conduct cross-scenario experiments where models are trained with one missing-modality configuration at a $70\%$ missing ratio and tested on different configurations. This evaluation is crucial for real-world deployment where the missing-modality patterns during inference may differ from those seen during training. We consider two training strategies: (1) training on both-missing scenarios where samples randomly have either text or image modality missing, and (2) training on single-modality scenarios where only one specific modality is consistently missing. We then evaluate these models on three test configurations: both-missing, image-missing, and text-missing scenarios.

As shown in \Cref{fig:generalizability_hatememes}, MoRA demonstrates superior cross-scenario generalization compared to DCP across all configurations on the Hateful Memes dataset. When trained on both-missing scenarios, MoRA maintains strong performance when tested on specific missing-modality cases, significantly outperforming DCP. This advantage persists even in challenging out-of-distribution scenarios—when models trained on text-missing data are tested on image-missing cases. The consistent performance gaps across all train-test combinations demonstrate that MoRA's dual mechanism of maintaining modality-specific parameters while enabling cross-modal knowledge transfer through Gram matrices creates a more robust representation space, particularly valuable for real-world deployments where missing-modality patterns may vary unpredictably from training conditions.

\subsection{Direction Property in MoRA}

To quantitatively validate our motivation illustrated in \Cref{fig:motivation}(c), we conduct comprehensive analysis comparing the embedding space of different approaches. We train models with $70\%$ missing ratio where both modalities are absent, then evaluate on complete test samples to measure how well each method preserves inter-modal relationships while enabling adaptation.

\textbf{Inter-modal Relationship Preservation.} We measure the average $L_2$ distance and angle between vision and text embeddings for each category in the Food101 test set. As shown in \Cref{tab:inter_modal}, MoRA maintains the inter-modal distance and alignment with the original CLIP. In contrast, FFT, which fine-tunes all parameters, severely distorts these relationships, while DCP shows substantial degradation. This demonstrates that MoRA successfully preserves the aligned embedding structure crucial for handling missing modalities.

\textbf{Modality-Specific Adaptation.} We analyze the embedding drift from original CLIP representations to measure modality-specific flexibility. \Cref{tab:drift} shows that MoRA achieves balanced adaptation with minimal drift, significantly outperforming FFT which exhibits catastrophic drift. DCP shows moderate drift but fails to maintain the inter-modal alignment as shown above.

\input{tables/tab_exp_for_motivation}
\input{tables/tab_peft_align_fusion}

\input{figures/fig_gram_eigen}
\subsection{Eigenvalue Analysis of MoRA}
We conduct an eigenspectrum analysis of the Gram matrices used in MoRA and compare them to the pre-trained weights. We extract eigenvalues from the Gram matrices and singular values from the pre-trained weights in layers $10$ and $11$ of the vision and text encoders. \Cref{fig:eigen} presents the normalized eigenvalue distributions. Our analysis reveals several critical findings. 
 
\textbf{First, Gram matrices serve as information concentration mechanisms}. This substantial difference demonstrates that Gram matrices effectively concentrate information in a much more compact form. To verify this empirically, we remove the Gram matrix, denoted as \texttt{w/o Gram} in \Cref{tab:align_fusion}. The average performance drops by $2.66\%$, confirming that Gram matrices are essential for effective knowledge transfer between modalities due to their information concentration properties.

\textbf{Second, text and vision modalities exhibit similar structural patterns in their Gram matrices}. Despite dimensional differences, we observe relatively stable eigenvalue distributions across indices for both modalities, indicating cross-modal structural similarities despite their dimensional differences. This structural similarity enables effective cross-modal knowledge transfer through the shared parameter space. To validate this, we use independent parameters instead of shared ones, denoted as \texttt{w/o Shared} in \Cref{tab:align_fusion}. The average performance drops by $2.78\%$, demonstrating that the emergent similar patterns are functionally critical for effective knowledge sharing.

\textbf{Third, we observe converging representational structures in deeper layers}. The eigenvalue pattern of Layer $11$ shows more convergence compared to Layer $10$, suggesting that deeper layers develop more aligned representational structures, which MoRA effectively leverages and maintains information preservation while enabling cross-modal transfer.

\input{tables/tab_cir}
\input{figures/fig_backbone_nparams_inference}
\subsection{Ablation Studies}
\textbf{Compared to Parameter-Efficient Fine-Tuning Methods}\quad We compare other parameter-efficient fine-tuning techniques, including LoRA~\citep{LoRA_ICLR22}, DoRA~\citep{DoRA_ICML24}, and BitFit~\citep{bitfit_acl22}, as shown in \Cref{tab:peft}. Low-rank-based methods achieve the best performance due to their flexibility. MoRA outperforms other methods, demonstrating its effectiveness in enabling modality interaction.

\textbf{Alternatives for Addressing Dimension Mismatch}\quad As shown in \Cref{tab:align_fusion}, \texttt{Align} uses two extra linear layers to project modalities into the same embedding, while \texttt{Fusion} concatenates embeddings from different modalities and uses one linear layer to project. MoRA consistently outperforms them across all datasets. We also conduct experiments removing modality-specific parameters in MoRA, denoted as \texttt{w/o Specific}. For Gram matrix construction, we report the performance of removing it and replacing it with a learnable one. These results validate the effectiveness of MoRA. We also add the ignored $\mathbf{W}_0$ in \Cref{eq:mora_ori}, showing that the learnable magnitude parameters effectively compensate for the omitted frozen weights during training.

Parameter sensitivity analysis can be found in \Cref{sec:sensitivity}.

\subsection{Extension to Embedding Tasks}

To demonstrate MoRA's generalizability beyond classification tasks, we evaluate it on Composed Image Retrieval (CIR) based on \citet{CIR_TOMM24} using CLIP models, where models use a reference image and text modification to identify target images. CIR models are trained on multimodal inputs, making the original image-to-image retrieval as an important and natural missing-modality scenario where texts are absent. We train models on the CIRR dataset~\citep{CIRR_ICCV21} with complete image-text pairs and evaluate on the MS-COCO validation set~\citep{MSCOCO_ECCV14} for image-to-image retrieval, as shown in \Cref{tab:cir}. MoRA achieves substantial improvements across all recall metrics, indicating that MoRA's applicability to tasks beyond classification, where missing modalities fundamentally alter the task dynamics. Implementation details can be found in \Cref{sec:implementation_details}.

\subsection{Scalability and Inference Time}
\Cref{fig:backbone} demonstrates the effectiveness of MoRA integration across various backbone architectures, including SLIP ViT-S~\citep{slip_eccv22}, CLIP ViT-B, and CLIP ViT-L~\citep{CLIP_ICML21}. The results indicate that performance exhibits favorable scaling properties with respect to model capacity, with accuracy improvements correlating positively with the number of parameters. We conduct a comprehensive analysis of inference times to evaluate the computational efficiency of MoRA and prompt-based methods, including MMP and DCP. As shown in \Cref{fig:nparams_inference}, prompt-based methods significantly increase the inference time. MoRA theoretically introduces no inference overhead, and experimental results demonstrate its efficiency.

%% file: tables/tab_main.tex
\begin{table}[t]
\caption{Performance comparison on MM-IMDb, Food101, and Hateful Memes datasets with varying missing ratios. MoRA consistently outperforms all baselines with average improvements of $5.30\%$, $1.91\%$, and $8.51\%$ over the next best method DCP.} \label{tab:main}
\centering
\resizebox{1.0\linewidth}{!}{
\begin{tabular}{c|c|cc|cccccc}
\toprule
Datasets & $\eta$ & Image & Text & CoOp & MMP & MaPLe & DePT & DCP & MoRA \\ \midrule
\multirow{10}{*}{MM-IMDb} 
& \multirow{4}{*}{50\%} 
& 100\% & 50\%  & 48.06 & 48.88 & 49.58 & 50.64 & 52.13 & \textbf{54.62 (+2.49)} \\
&                       & 50\%  & 100\% & 49.89 & 51.46 & 52.32 & 52.78 & 54.32 & \textbf{57.61 (+3.29)} \\
&                       & 75\%  & 75\%  & 48.37 & 49.32 & 49.56 & 50.87 & 52.32 & \textbf{55.88 (+3.56)} \\
& & \multicolumn{2}{c|}{\cellcolor{gray!15}\textit{Average}} & \cellcolor{gray!15}48.77 & \cellcolor{gray!15}49.89 & \cellcolor{gray!15}50.49 & \cellcolor{gray!15}51.43 & \cellcolor{gray!15}52.92 & \cellcolor{gray!15}\textbf{56.04 (+3.12)} \\ 
\cmidrule(l){2-10} 
& \multirow{4}{*}{70\%} 
& 100\% & 30\%  & 44.13 & 45.64 & 45.52 & 46.38 & 48.52 & \textbf{52.56 (+4.04)} \\
&                       & 30\%  & 100\% & 48.82 & 50.52 & 50.64 & 52.13 & 53.14 & \textbf{56.39 (+3.25)} \\
&                       & 65\%  & 65\%  & 46.84 & 48.12 & 49.16 & 50.32 & 51.42 & \textbf{52.97 (+1.55)} \\ 
& & \multicolumn{2}{c|}{\cellcolor{gray!15}\textit{Average}} & \cellcolor{gray!15}46.60 & \cellcolor{gray!15}48.09 & \cellcolor{gray!15}48.44 & \cellcolor{gray!15}49.61 & \cellcolor{gray!15}51.03 & \cellcolor{gray!15}\textbf{53.97 (+2.94)} \\
\cmidrule(l){2-10}
& \multirow{4}{*}{90\%} 
& 100\% & 10\%  & 44.76 & 45.32 & 46.84 & 47.56 & 49.26 & \textbf{50.67 (+1.41)} \\
&                       & 10\%  & 100\% & 48.32 & 49.12 & 50.13 & 50.88 & 52.22 & \textbf{53.57 (+1.35)} \\
&                       & 55\%  & 55\%  & 44.12 & 44.87 & 45.12 & 46.54 & 48.04 & \textbf{51.64 (+3.60)} \\ 
& & \multicolumn{2}{c|}{\cellcolor{gray!15}\textit{Average}} & \cellcolor{gray!15}45.73 & \cellcolor{gray!15}46.44 & \cellcolor{gray!15}47.36 & \cellcolor{gray!15}48.33 & \cellcolor{gray!15}49.84 & \cellcolor{gray!15}\textbf{51.96 (+2.12)} \\
\midrule
\multirow{10}{*}{Food101} 
& \multirow{4}{*}{50\%} 
& 100\% & 50\%  & 77.45 & 77.89 & 79.64 & 80.16 & 82.11 & \textbf{84.41 (+2.30)} \\
&                       & 50\%  & 100\% & 87.02 & 87.16 & 87.35 & 82.14 & 89.12 & \textbf{89.63 (+0.51)} \\
&                       & 75\%  & 75\%  & 81.24 & 81.72 & 82.34 & 83.12 & 85.24 & \textbf{86.68 (+1.44)} \\ 
& & \multicolumn{2}{c|}{\cellcolor{gray!15}\textit{Average}} & \cellcolor{gray!15}81.90 & \cellcolor{gray!15}82.26 & \cellcolor{gray!15}83.11 & \cellcolor{gray!15}81.81 & \cellcolor{gray!15}85.49 & \cellcolor{gray!15}\textbf{86.91 (+1.42)} \\
\cmidrule(l){2-10}
& \multirow{4}{*}{70\%} 
& 100\% & 30\%  & 76.34 & 76.52 & 77.02 & 77.34 & 78.87 & \textbf{80.85 (+1.98)} \\
&                       & 30\%  & 100\% & 84.78 & 85.64 & 85.89 & 86.12 & 87.32 & \textbf{88.01 (+0.69)} \\
&                       & 65\%  & 65\%  & 78.87 & 79.12 & 79.84 & 81.46 & 81.87 & \textbf{83.77 (+1.90)} \\ 
& & \multicolumn{2}{c|}{\cellcolor{gray!15}\textit{Average}} & \cellcolor{gray!15}80.00 & \cellcolor{gray!15}80.43 & \cellcolor{gray!15}80.92 & \cellcolor{gray!15}81.64 & \cellcolor{gray!15}82.69 & \cellcolor{gray!15}\textbf{84.21 (+1.52)} \\
\cmidrule(l){2-10}
& \multirow{4}{*}{90\%} 
& 100\% & 10\%  & 71.87 & 73.14 & 73.46 & 74.12 & 75.26 & \textbf{78.41 (+3.15)} \\
&                       & 10\%  & 100\% & 81.67 & 82.14 & 83.12 & 83.56 & 85.78 & \textbf{86.77 (+0.99)} \\
&                       & 55\%  & 55\%  & 76.46 & 76.58 & 77.85 & 78.12 & 79.87 & \textbf{81.09 (+1.22)} \\ 
& & \multicolumn{2}{c|}{\cellcolor{gray!15}\textit{Average}} & \cellcolor{gray!15}76.67 & \cellcolor{gray!15}77.29 & \cellcolor{gray!15}78.14 & \cellcolor{gray!15}78.60 & \cellcolor{gray!15}80.30 & \cellcolor{gray!15}\textbf{82.09 (+1.79)} \\
\midrule
\multirow{10}{*}{Hateful Memes} 
& \multirow{4}{*}{50\%} 
& 100\% & 50\%  & 60.56 & 60.31 & 60.87 & 61.87 & 62.32 & \textbf{70.66 (+8.34)} \\
&                       & 50\%  & 100\% & 62.41 & 62.35 & 63.13 & 63.88 & 64.46 & \textbf{71.58 (+7.12)} \\
&                       & 75\%  & 75\%  & 64.87 & 65.84 & 65.46 & 65.86 & 66.02 & \textbf{69.58 (+3.56)} \\ 
& & \multicolumn{2}{c|}{\cellcolor{gray!15}\textit{Average}} & \cellcolor{gray!15}62.61 & \cellcolor{gray!15}62.83 & \cellcolor{gray!15}63.15 & \cellcolor{gray!15}63.87 & \cellcolor{gray!15}64.27 & \cellcolor{gray!15}\textbf{70.61 (+6.34)} \\
\cmidrule(l){2-10}
& \multirow{4}{*}{70\%} 
& 100\% & 30\%  & 60.74 & 61.12 & 61.26 & 61.56 & 62.82 & \textbf{69.43 (+6.61)} \\
&                       & 30\%  & 100\% & 62.74 & 63.24 & 63.14 & 63.48 & 64.12 & \textbf{70.68 (+6.56)} \\
&                       & 65\%  & 65\%  & 64.82 & 65.04 & 65.23 & 65.48 & 66.08 & \textbf{70.15 (+4.07)} \\ 
& & \multicolumn{2}{c|}{\cellcolor{gray!15}\textit{Average}} & \cellcolor{gray!15}62.77 & \cellcolor{gray!15}63.13 & \cellcolor{gray!15}63.21 & \cellcolor{gray!15}63.51 & \cellcolor{gray!15}64.34 & \cellcolor{gray!15}\textbf{70.09 (+5.75)} \\
\cmidrule(l){2-10}
& \multirow{4}{*}{90\%} 
& 100\% & 10\%  & 60.03 & 57.21 & 60.74 & 61.14 & 62.08 & \textbf{68.52 (+6.44)} \\
&                       & 10\%  & 100\% & 61.46 & 61.52 & 61.87 & 62.42 & 63.87 & \textbf{68.78 (+4.91)} \\
&                       & 55\%  & 55\%  & 64.32 & 63.34 & 64.85 & 65.37 & 66.78 & \textbf{68.37 (+1.59)} \\ 
& & \multicolumn{2}{c|}{\cellcolor{gray!15}\textit{Average}} & \cellcolor{gray!15}61.94 & \cellcolor{gray!15}60.69 & \cellcolor{gray!15}62.49 & \cellcolor{gray!15}62.98 & \cellcolor{gray!15}64.24 & \cellcolor{gray!15}\textbf{68.56 (+4.32)} \\
\bottomrule
\end{tabular}
}
\end{table}

%% file: figures/fig_generalizability_hatememes.tex
\begin{figure}[t]
    \centering
    \includegraphics[width=0.95\linewidth]{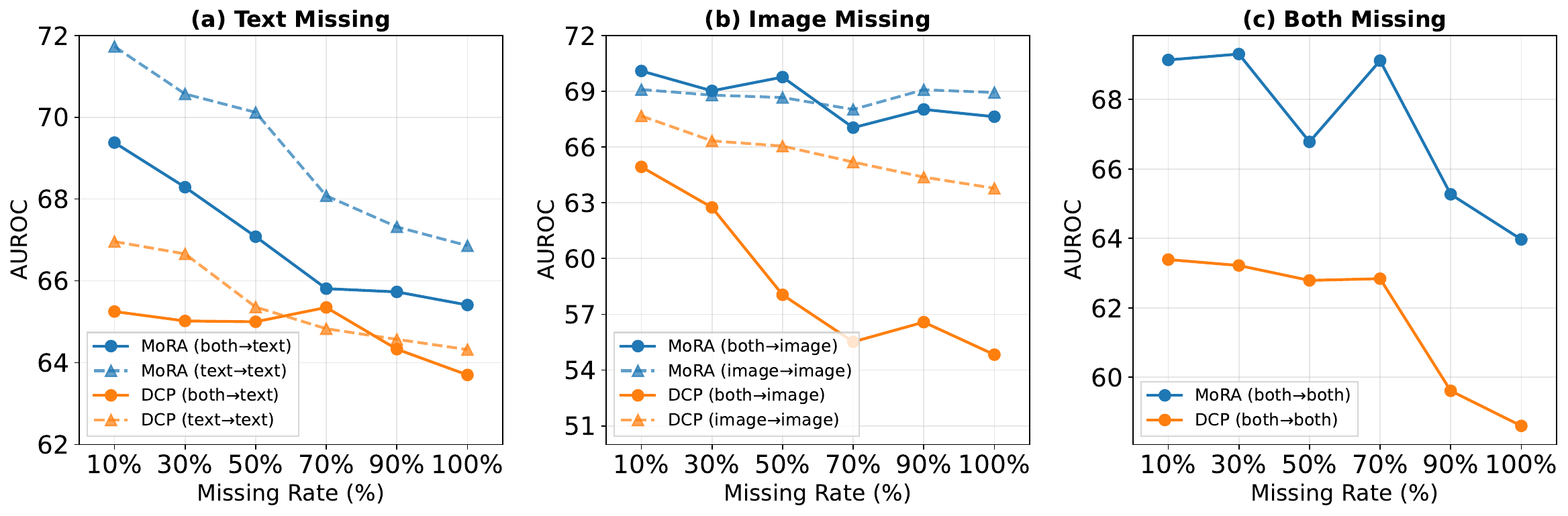}
    \caption{Generalizability Analysis on Hateful Memes dataset. (a) Models are trained on missing-both or missing-text cases, and evaluated on missing-text cases. (b) Models are trained on missing-both or missing-image cases, and evaluated on missing-image cases. (c) All models are trained on missing-both cases, and evaluated on missing-both cases.}
    \label{fig:generalizability_hatememes}
\end{figure}

%% file: tables/tab_exp_for_motivation.tex
\begin{table}[t]
\centering
\begin{minipage}{0.48\textwidth}
\centering
\captionof{table}{Inter-modal Distance Analysis. Average $L_2$ distance and angle between vision and text embeddings on Food101 test set.}
\label{tab:inter_modal}
\resizebox{0.8\linewidth}{!}{
\begin{tabular}{lrr}
\toprule
Method & $L_2$ Dist. & Angle ($^\circ$) \\
\midrule
CLIP (orig.) & 1.18 & 72.44 \\
FFT & 22.61 & 91.64 \\
DCP & 15.78 & 86.92 \\
\cellcolor{gray!15}\textbf{MoRA} & \cellcolor{gray!15}\textbf{9.99} & \cellcolor{gray!15}\textbf{77.07} \\
\bottomrule
\end{tabular}
}
\end{minipage}
\hfill
\begin{minipage}{0.48\textwidth}
\centering
\captionof{table}{Modality-Specific Drift Analysis. Average embedding shift from original CLIP representations.}
\label{tab:drift}
\resizebox{0.8\linewidth}{!}{
\begin{tabular}{lrrrr}
\toprule
\multirow{2}{*}{Method} & \multicolumn{2}{c}{Vision} & \multicolumn{2}{c}{Text} \\
\cmidrule(lr){2-3} \cmidrule(lr){4-5}
& $L_2$ & Angle & $L_2$ & Angle \\
\midrule
FFT & 8.36 & 92.16 & 20.60 & 87.67 \\
DCP & 8.22 & 65.17 & 13.57 & 66.26 \\
\cellcolor{gray!15}\textbf{MoRA} & \cellcolor{gray!15}\textbf{8.12} & \cellcolor{gray!15}\textbf{43.24} & \cellcolor{gray!15}\textbf{6.04} & \cellcolor{gray!15}\textbf{44.84} \\
\bottomrule
\end{tabular}
}
\end{minipage}
\end{table}

%% file: tables/tab_peft_align_fusion.tex
\begin{table}[t]
\begin{minipage}{0.43\textwidth}
    \centering
    \captionof{table}{Performance comparison of different parameter-efficient fine-tuning methods.}
    \label{tab:peft}
    \resizebox{1\linewidth}{!}{
    \begin{tabular}{lccc}
    \toprule
    & MM-IMDb & Food101 & Hateful Memes \\
    \midrule
    \rowcolor{gray!15} 
    \multicolumn{4}{l}{\textit{Low-Rank-Based}} \\
    MoRA & \textbf{52.97} & \textbf{83.77} & \textbf{70.15} \\
    LoRA & 51.35 & 82.14 & 67.97 \\
    DoRA & 51.89 & 82.34 & 68.28 \\
    \rowcolor{gray!15} 
    \multicolumn{4}{l}{\textit{Prompt-Based}} \\
    DePT & 50.32 & 81.46 & 65.48 \\
    DCP & 51.42 & 81.87 & 66.08 \\
    \rowcolor{gray!15} 
    \multicolumn{4}{l}{\textit{Weight Fine-Tuning}} \\
    BitFit & 48.57 & 79.38 & 64.10 \\
    FFT & 3.01 & 14.05 & 46.91 \\
    \bottomrule
    \end{tabular}
    }
\end{minipage}%
\hfill
\begin{minipage}{0.52\textwidth}
    \centering
    \captionof{table}{Comparison with multimodal alignment and fusion methods.}
    \label{tab:align_fusion}
    \resizebox{1\linewidth}{!}{
    \begin{tabular}{lccc}
    \toprule
        & MM-IMDb & Food101 & Hateful Memes \\ \midrule
    \cellcolor{gray!15}MoRA & \cellcolor{gray!15}\textbf{52.97} & \cellcolor{gray!15}\textbf{83.77} & \cellcolor{gray!15}\textbf{70.15} \\
    Align & 51.39 & 81.14 & 68.53 \\
    Fusion & 50.72 & 81.01 & 68.17 \\ 
    w/o Specific & 51.18 & 81.32 & 68.71\\
    w/o Gram & 50.41 & 80.31 & 68.19 \\
    \midrule
    w/ Learnable Gram & 52.25 & 83.37 & 69.12 \\
    w/ $\mathbf{W}_0$ & 52.88 & 83.59 & 70.03 \\
    \bottomrule
    \end{tabular}
    }
\end{minipage}
\end{table}

%% file: figures/fig_gram_eigen.tex
\begin{figure}[t]
    \centering
    \includegraphics[width=0.8\linewidth]{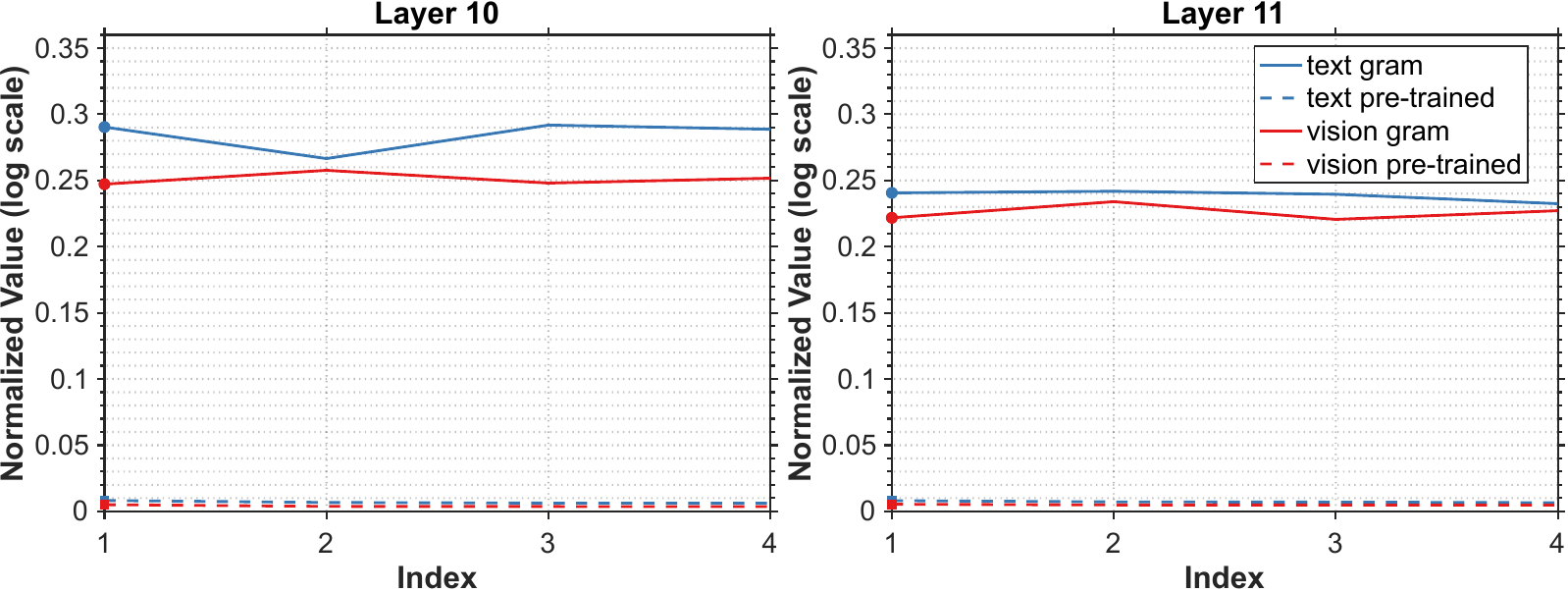}
    \caption{Comparison of eigenvalue distributions between Gram matrices and pre-trained weights.}
    \label{fig:eigen}
\end{figure}

%% file: tables/tab_cir.tex
\begin{table}[t]
\centering
\caption{Performance comparison on image-to-image retrieval using models trained on multimodal CIRR dataset. Results are evaluated on the MS-COCO validation set.}
\label{tab:cir}
\resizebox{0.7\linewidth}{!}{
\begin{tabular}{lccc}
\toprule
Method & Recall@1 & Recall@5 & Recall@10 \\
\midrule
CLIP4CIR~\citep{CIR_TOMM24} & 43.34 & 76.99 & 86.49 \\
\cellcolor{gray!15}CLIP4CIR + MoRA & \cellcolor{gray!15}\textbf{60.50} & \cellcolor{gray!15}\textbf{85.00} & \cellcolor{gray!15}\textbf{88.60} \\
\bottomrule
\end{tabular}
}
\end{table}

%% file: figures/fig_backbone_nparams_inference.tex
\begin{figure}[t]
\begin{minipage}{0.48\textwidth}
    \centering
    \includegraphics[width=0.9\linewidth]{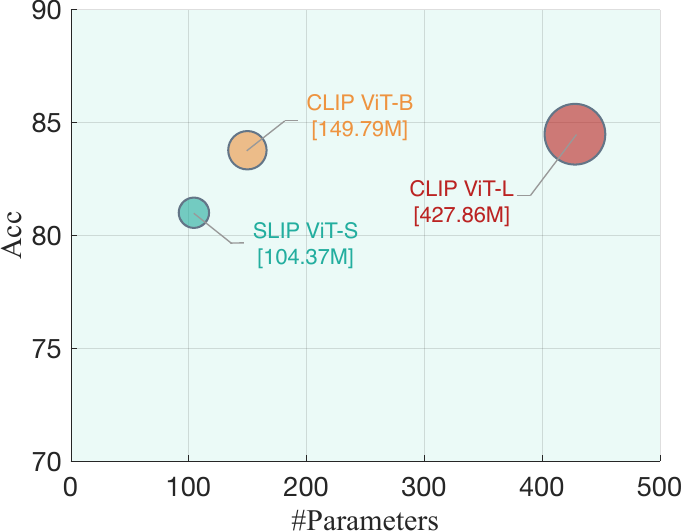}
    \captionof{figure}{Performance scaling of MoRA with different backbone models.}
    \label{fig:backbone}
\end{minipage}%
\hfill
\begin{minipage}{0.48\textwidth}
    \centering
    \includegraphics[width=0.9\linewidth]{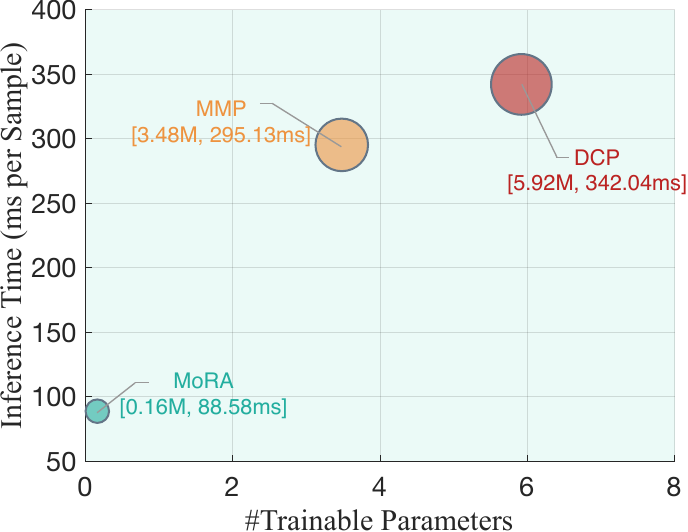}
    \captionof{figure}{Inference time (ms) per sample versus the number of trainable parameters.}
    \label{fig:nparams_inference}
\end{minipage}
\end{figure}

%% file: sections/5_conclusion.tex
\section{Conclusion}
We introduced MoRA, a parameter-efficient fine-tuning method that effectively addresses the missing modality challenge in multimodal learning through shared cross-modal parameters and modality-specific adaptations. By leveraging Gram matrices for dimension-agnostic knowledge transfer, MoRA enables bidirectional information exchange while preserving modality-specific characteristics without introducing inference overhead. Extensive experiments demonstrate that MoRA significantly outperforms existing approaches across multiple benchmarks both on performance and inference time, demonstrating the effectiveness and efficiency of MoRA.

%% file: sections/appendix.tex
\newpage
\appendix
\section{Details of Experimental Setup} \label{sec:experiment_setup_details}
\subsection{Dataset}
We evaluate our proposed method on three standard benchmarks: MM-IMDb~\citep{mmimdb_iclrw17}, UPMC-Food101~\citep{food101_icmew15}, and Hateful Memes~\citep{hatefulmemes_neurips20}.

\textbf{MM-IMDb} represents the largest publicly available multimodal collection for movie genre prediction, containing $25{,}959$ movies annotated with both visual and textual information. This dataset supports multi-label classification across $27$ distinct movie genres. The corpus is structured with $15{,}552$ training, $2{,}608$ validation, and $7{,}799$ test image-text pairs, providing a robust foundation for developing and evaluating multimodal classification models. 

\textbf{UPMC Food101} is a comprehensive multimedia collection featuring noisy image-text pairs gathered from Google Image Search across $101$ food categories. The dataset is structured with $61{,}127$ training samples, $6{,}845$ validation samples, and $22{,}716$ test image-text pairs, providing substantial material for developing and evaluating multimodal food recognition systems. 

\textbf{Hateful Memes} represents a benchmark multimodal collection for detecting hate speech in memes. It contains over $10{,}000$ image-text pairs specifically designed to evaluate multimodal reasoning capabilities, where the interplay between text and visuals is crucial for accurate classification. The dataset consists of $8{,}500$, $500$, and $1{,}000$ samples for training, validation, and testing.

\subsection{Baseline Methods}
To evaluate MoRA, we select the SOTA missing-modality methods and multimodal prompt methods. Specifically, we select the missing modality methods, including MMP~\citep{MMP_CVPR23} and DCP~\citep{DCP_NeurIPS24}; for multimodal prompt learning, we choose CoOp~\citep{coop_ijcv22}, MaPLe~\citep{MaPLe_CVPR23}, and DePT~\citep{dept_cvpr24}. Although a recent work, SyP~\citep{SYP_arXiv25}, employs the prompt-based method to address the missing modality task, the code for this work was not released upon our submission. Therefore, we do not compare our method with it. Once the source code or pre-trained models are released, we will add the results to the main results.

\subsection{Implementation Details} \label{sec:implementation_details}
\paragraph{Main Experiments} Following previous work~\citep{DCP_NeurIPS24}, we use CLIP ViT-B/16 as the backbone model. We add a fully connected layer at the top of the model as the classification layer for downstream tasks. The parameters in the CLIP model are frozen, and we only fine-tune the parameters of the classification layer and MoRA modules. MoRA can be inserted into various positions in the backbone. We found that the best hyper-parameters differ in various datasets. In UPMC-Food101, MoRA is inserted into the $\mathbf{Q},\mathbf{V}$ in self-attention modules of the last two vision and text transformer layers. Rank $r$ is $4$. We use the AdamW~\citep{adamw_iclr19} optimizer with a learning rate $0.01$ and weight decay $0.02$. A linear warmup cosine annealing scheduler with $10\%$ warmup steps is used to adjust the learning rate. The batch size is $256$. The number of training epochs is $20$ and we apply the early-stopping strategy. Detailed settings can be found in the code we provided. If not specified, experiments are conducted on the UPMC-Food101 dataset with a missing ratio $\eta$ of $70\%$, where both image and text modalities are absent. We run experiments three times and report their average performance. All experiments are conducted on one NVIDIA H100 GPU.

\paragraph{Motivation} In \Cref{fig:motivation}, most hyper-parameters are the same as those used in the main experiments. We use CLIP ViT-B/16 as the aligned model and pre-trained ViT-B/16 and BERT as the unaligned model. We train these models on complete training datasets, i.e., $\eta=0\%$, and test them using different datasets, including complete, image, and text-only datasets.

\paragraph{Embedding Task} In \Cref{tab:cir}, we use the same settings in the main experiments above. We use the CIRR training data as the training set, and evaluate the trained model on the MS-COCO validation set. We select CLIP ViT-B/16 as the backbone. For training, we use the complete samples without any modality-incomplete data. During testing, the model is evaluated on the image-to-image retrieval task, which can be viewed as if the text modality is missing.

\section{Attached Position}
We systematically evaluate MoRA attachment at different network depths, as shown in \Cref{fig:sensitivity}. To further analyze the effect of attached positions, we attached MoRA to three positions of a CLIP ViT-B/16 model:
\begin{itemize}
    \item Front layers: Layers 1-2 (early feature extraction)
    \item Middle layers: Layers 6-7 (intermediate representations)
    \item Rear layers: Layers 11-12 (high-level semantics)
\end{itemize}

We train the model on the Food101 dataset with a $70\%$ missing ratio and both modalities are missing. As shown in \Cref{tab:attach_position}, attaching to deeper layers enables fine-tuning of high-level semantic features rather than low-level representations, yielding superior performance. This strategy effectively handles architectures with asymmetric depths. As demonstrated in \Cref{fig:backbone}, MoRA successfully adapts CLIP ViT-L (24 vision layers, 12 text layers) by consistently targeting the final layers of each modality, which contain the most semantic information.

\input{figures/fig_generalizability_mmimdb}
\input{figures/fig_generalizability_food101}
\input{tables/tab_attach_position}
\section{More Results} \label{sec:more_results}
More results across various missing ratios are shown in \Cref{fig:generalizability_mmimdb} and \Cref{fig:generalizability_food101}. The experimental results demonstrate consistent effectiveness in handling missing modalities. As the missing ratio increases, performance on all datasets gradually declines. Notably, the text-only modality consistently outperformed image-only across all datasets. MoRA maintains robust performance even at high missing ratios, preserving inter-modality interactions while maintaining intra-modality flexibility.

\section{Parameter Sensitivity} \label{sec:sensitivity}
\input{figures/fig_sensitivity}

The sensitivity of MoRA to its key hyper-parameters is shown in \Cref{fig:sensitivity}, where $r$ denotes the rank, $\eta$ is the low-rank strength, ``\#Layers'' denotes the number of layers inserted by MoRA, and ``Position'' means which attention matrices are adjusted. The results show that MoRA is robust to parameter changes, maintaining strong performance across a wide range of values.

\section{Visualization of Embedding Space}
\input{figures/fig_tsne}
To further analyze why MoRA outperforms other methods, we use t-SNE~\citep{tsne_jmlr08} to visualize the embeddings of samples with missing modalities, as illustrated in \Cref{fig:tsne}. Specifically, we use the samples in the test dataset, obtain the embeddings from available modalities, and visualize them. The results show that the embedding space of \texttt{FFT} has collapsed, and MoRA produces more compact and well-separated clusters. Compared to \texttt{DCP}, MoRA has a larger inter-class distance, indicating better discriminability.

\section{LLM Usage Statement}

Large language models were used as a general-purpose writing assistance tool during the preparation of this manuscript, primarily for grammar checking, sentence restructuring, and improving clarity of technical descriptions. LLMs did not contribute to the core research ideas, experimental design, or technical innovations presented in this work. All scientific claims, experimental results, and theoretical contributions are the original work of the authors, who take full responsibility for the accuracy and integrity of all content.

\section{Limitations} \label{sec:limitations}
While MoRA demonstrates strong performance across various missing-modality scenarios, several limitations present opportunities for future research.

First, our experimental validation is limited to three datasets (MM-IMDb, UPMC-Food101, and Hateful Memes) and primarily focuses on image-text modality pairs. Future work could extend MoRA to additional multimodal domains (e.g., audio-visual) and more diverse datasets to further validate its generalizability. Second, the current formulation of MoRA addresses binary missing-modality scenarios (present or absent). Future work could explore extensions to partial or corrupted modalities where information is present but degraded, which may better reflect certain real-world applications.

Despite these limitations, MoRA represents a significant step forward in addressing the missing modality challenge through its novel parameter-efficient fine-tuning approach.

\section{Broader Impacts} \label{sec:broader_impacts}
Our research on MoRA offers several positive societal impacts. By addressing the missing modality challenge in multimodal systems, MoRA can significantly improve accessibility for users with sensory impairments who may not have access to all modalities. Additionally, MoRA reduces computational requirements and potentially lowers energy consumption compared to alternative methods, contributing to more sustainable AI development. This efficiency also enables more robust deployment of multimodal systems in resource-constrained environments like healthcare, education, and humanitarian assistance. Furthermore, MoRA could enhance privacy by allowing users to selectively withhold certain modalities while still receiving reasonable system performance.

We also acknowledge potential concerns regarding this technology. As with many AI advancements, improvements in handling missing modalities could potentially be applied in ways that raise privacy questions if deployed without appropriate safeguards. Additionally, systems making decisions based on incomplete information should be deployed with appropriate human oversight, particularly in high-stakes applications. We've focused our development on public benchmark datasets and emphasize that our primary goal is improving the accessibility, efficiency, and robustness of multimodal systems rather than enabling capabilities that could raise significant ethical concerns.

%% file: figures/fig_generalizability_mmimdb.tex
\begin{figure}[t]
    \centering
    \includegraphics[width=1.0\linewidth]{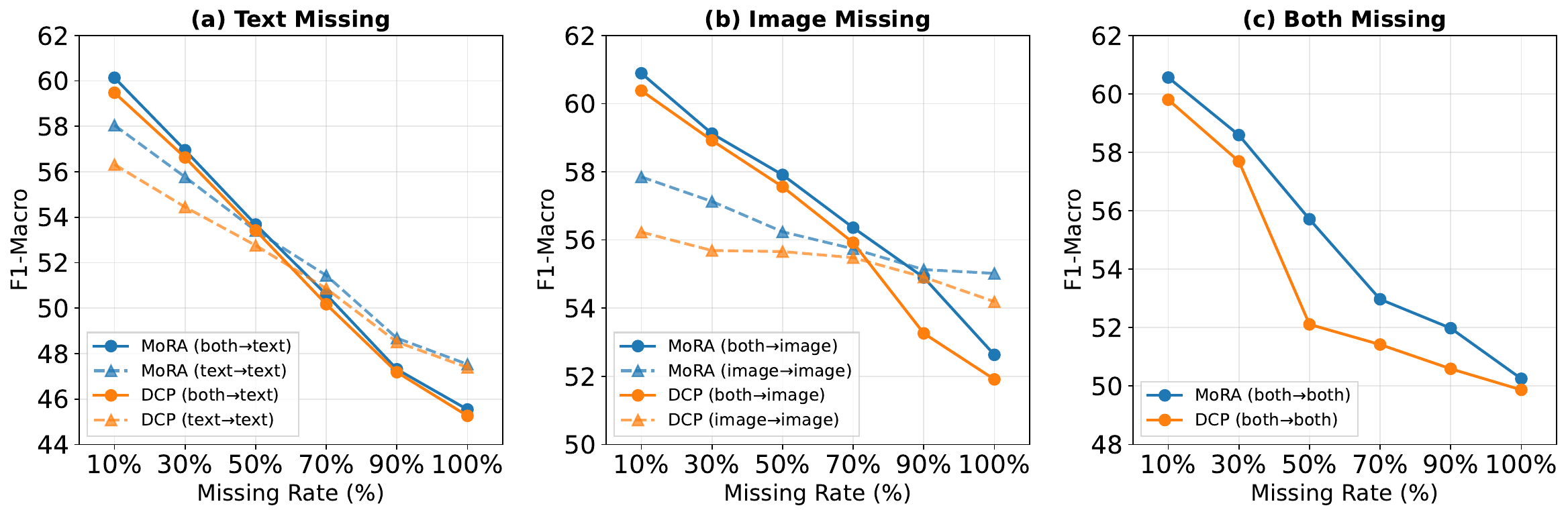}
    \caption{Performance comparison on MM-IMDb with varying missing ratios.}
    \label{fig:generalizability_mmimdb}
\end{figure}

%% file: figures/fig_generalizability_food101.tex
\begin{figure}[t]
    \centering
    \includegraphics[width=1.0\linewidth]{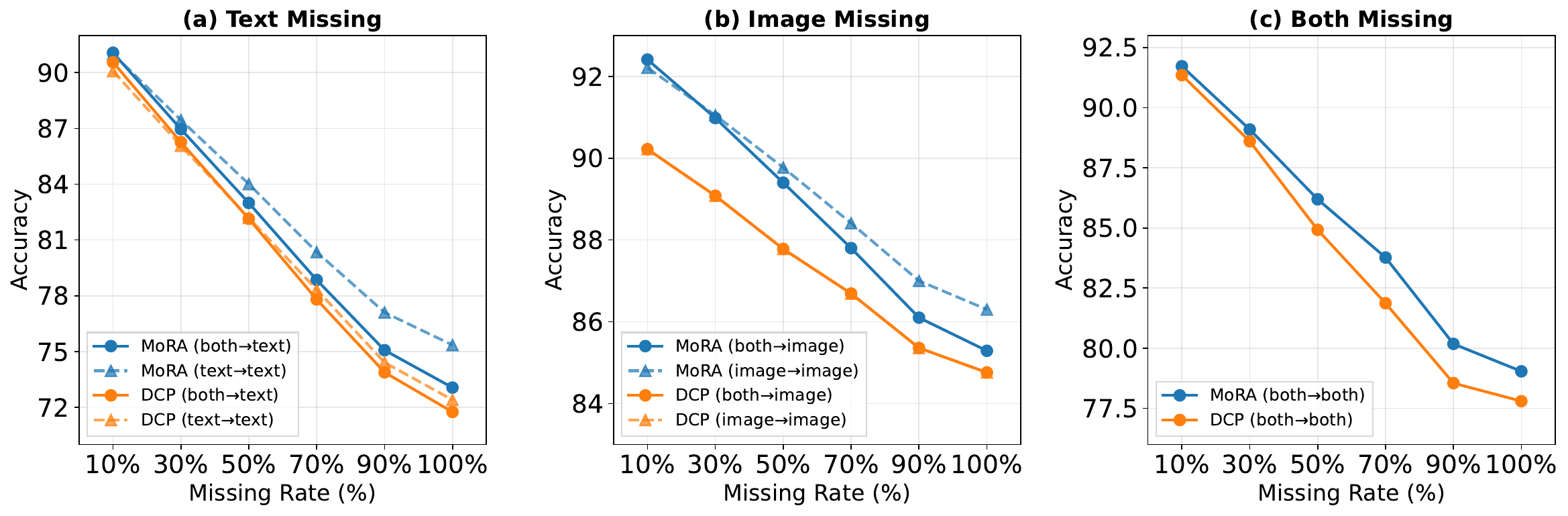}
    \caption{Performance comparison on Food101 with varying missing ratios.}
    \label{fig:generalizability_food101}
\end{figure}

%% file: tables/tab_attach_position.tex
\begin{table}[t]
    \centering
    \caption{Performance comparison of attached positions.}
    \label{tab:attach_position}
    
    \begin{tabular}{l|ccc}
    \toprule
    Position & Front & Middle & \textbf{Rear~(Ours)} \\
    \midrule
    Accuracy & 81.08 & 82.48 & \textbf{83.77} \\
    \bottomrule
    \end{tabular}
\end{table}

%% file: figures/fig_sensitivity.tex
\begin{figure}[t]
    \centering
    \includegraphics[width=0.90\linewidth]{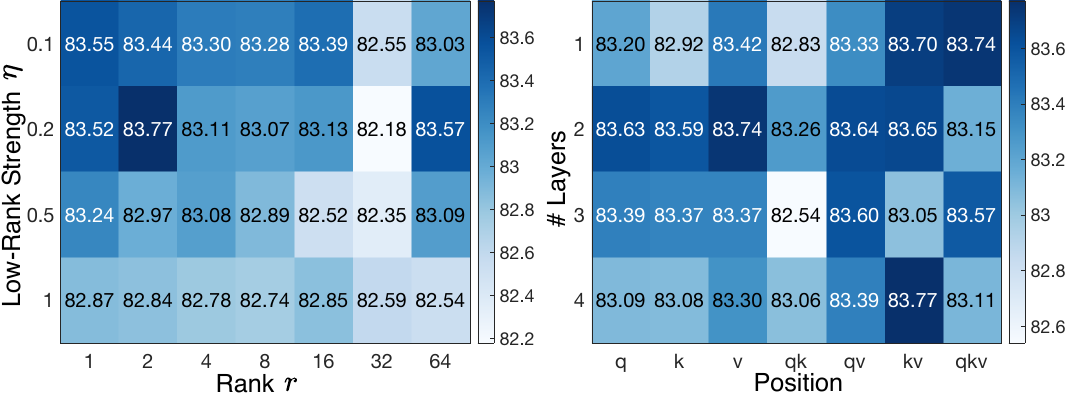}
    \caption{Parameter sensitivity analysis.}
    \label{fig:sensitivity}
\end{figure}

%% file: figures/fig_tsne.tex
\begin{figure}[t]
    \centering
    \includegraphics[width=0.9\linewidth]{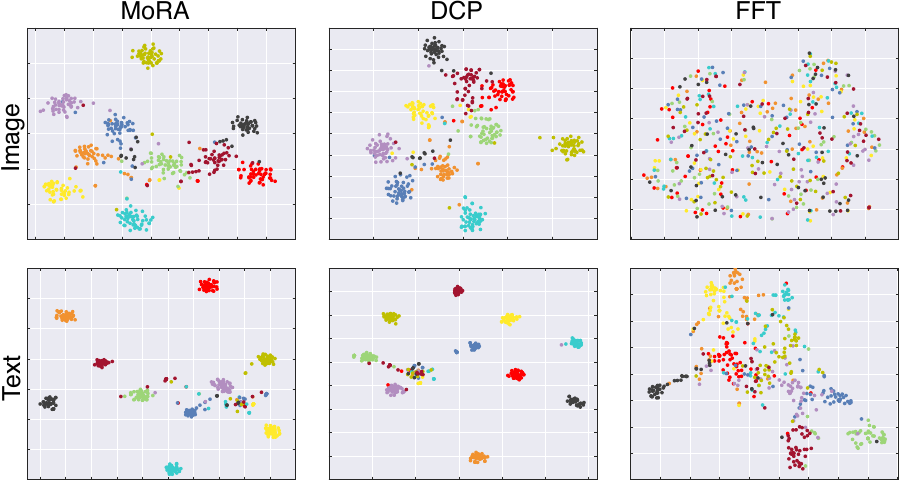}
    \caption{t-SNE visualization for MoRA, DCP, and FFT. Different colors denote different categories.}
    \label{fig:tsne}
\end{figure}